\documentclass[runningheads]{llncs}
\usepackage[T1]{fontenc}
\usepackage{amsmath}
\usepackage{amsfonts}
\usepackage{graphicx,verbatim}
\usepackage{booktabs}
\usepackage{multirow}
\usepackage{adjustbox}
\usepackage{hyperref}
\begin{document}
	\title{Prototype-Guided and Lightweight Adapters for Inherent Interpretation and Generalisation in Federated Learning}
	\titlerunning{Prototype-Guided and Adapters for Interpretable Federated Learning}
	\author{Samuel Ofosu Mensah\inst{1}\and
		Kerol Djoumessi\inst{1}\and
		Philipp Berens\inst{1,2}}
	\authorrunning{S.O. Mensah et al.}
	\institute{Hertie Institute for AI in Brain Health, University of Tübingen, Germany \and
		Tübingen AI Center, University of Tübingen, Germany\\
		\email{\{samuel.ofosu-mensah, philipp.berens\}@uni-tuebingen.de}
	}
	\maketitle
	\begin{abstract}
		Federated learning (FL) provides a promising paradigm for collaboratively training machine learning models across distributed data sources while maintaining privacy. Nevertheless, real-world FL often faces major challenges including communication overhead during the transfer of large model parameters and statistical heterogeneity, arising from non-identical independent data distributions across clients. In this work, we propose an FL framework that 1) provides inherent interpretations using prototypes, and 2) tackles statistical heterogeneity by utilising lightweight adapter modules to act as compressed surrogates of local models and guide clients to achieve generalisation despite varying client distribution. Each client locally refines its model by aligning class embeddings toward prototype representations and simultaneously adjust the lightweight adapter.  Our approach replaces the need to communicate entire model weights with prototypes and lightweight adapters. This design ensures that each client’s model aligns with a globally shared structure while minimising communication load and providing inherent interpretations. Moreover, we conducted our experiments on a real-world retinal fundus image dataset, which provides clinical-site information. We demonstrate inherent interpretable capabilities and perform a classification task, which shows improvements in accuracy over baseline algorithms\footnote{Code is available at 
			\url{https://github.com/berenslab/FedAdapterProto}
		}. 
	\end{abstract}
	\section{Introduction}
	Federated Learning (FL) has gained considerable attention due to its ability to collaboratively train machine learning models across distributed data sources while preserving privacy and reducing the need for data centralisation~\cite{li,mcmahan}. This is particularly relevant in healthcare, where patient privacy and data regulations often prevent direct data sharing between institutions~\cite{brisimi,rieke,Sadilek}. FL offers numerous advantages in the medical domain by providing privacy-preserving collaboration among medical research institutions~\cite{rieke},enabling the development of generalisable models using diverse patient data while reducing regulatory challenges. However, medical imaging data presents unique challenges in FL mainly due to statistical heterogeneity across institutions arising from differences in image acquisition protocols, diverse patient populations, imbalance class distributions, and other factors~\cite{guan,yang,yoo}. As a result, clients in FL often deal with non-identically independent distributed (non-IID) data, leading to inconsistent client updates and degradation of global model performance~\cite{gong,wu}.
	\\
	\\
	Since the introduction of FedAvg~\cite{mcmahan}, many FL algorithms have been proposed to address these challenges~\cite{cai,huang,li_2,liang,tan,wang}. Nevertheless, most methods communicate and aggregate full model parameters resulting in high communication costs. To address heterogeneity and communication constraints, one approach is to communicate representations of local models such as prototypes~\cite{snell} or compressed versions of the model~\cite{wu_c}. Communicating prototypes rather than full model parameters reduces communication overhead, encourages alignment of class representations across clients, and improves tolerance to non-IID data~\cite{tan}. However, prototypes do not adapt the feature extractor for each client. Also, it is important to mention that the definition of prototypes in FL varies across literature. Most studies define prototypes as the mean value of a backbone's feature map belonging to a specific class. While these definitions improves communication, they do not provide inherent interpretations. We adopt the definition of prototypes as described in~\cite{chen}. Another line of work leverages adapters -- a small learnable bottleneck -- to address representation misalignment caused by heterogeneity~\cite{cai}. While effective, these approaches do not provide a mechanism for model interpretability. In summary, existing methods either communicate full models or specialised representations, but trade off communication efficiency, personalisation, or interpretability. 
	\\
	\\
	To overcome these limitations, we proposed a novel FL framework that communicates only adapter modules~\cite{cai,chen_h} and prototypes~\cite{chen}. By exchanging only lightweight adapters and class-representative prototypes, our approach handles statistical heterogeneity with minimal communication overhead~\cite{cai,tan}.
	Additionally, the use of prototypes offers inherent interpretations into the model’s decisions. To the best of our knowledge, this is the first work to integrate adapter-based fine-tuning with prototype-based learning in FL. Our key contributions are to; 1) develop an FL method that communicates only adapter parameters and prototype parameters, mitigating non-IID discrepancies across clients and 2) provide an FL method that provides inherent interpretations -- a feature largely absent in prior FL methods. We carried out our experiments using a real-world retinal fundus image dataset and classified diabetic retinopathy (DR).
	\section{Methodology}
	\begin{figure}[t]
		\centering
		\includegraphics[width=\textwidth]{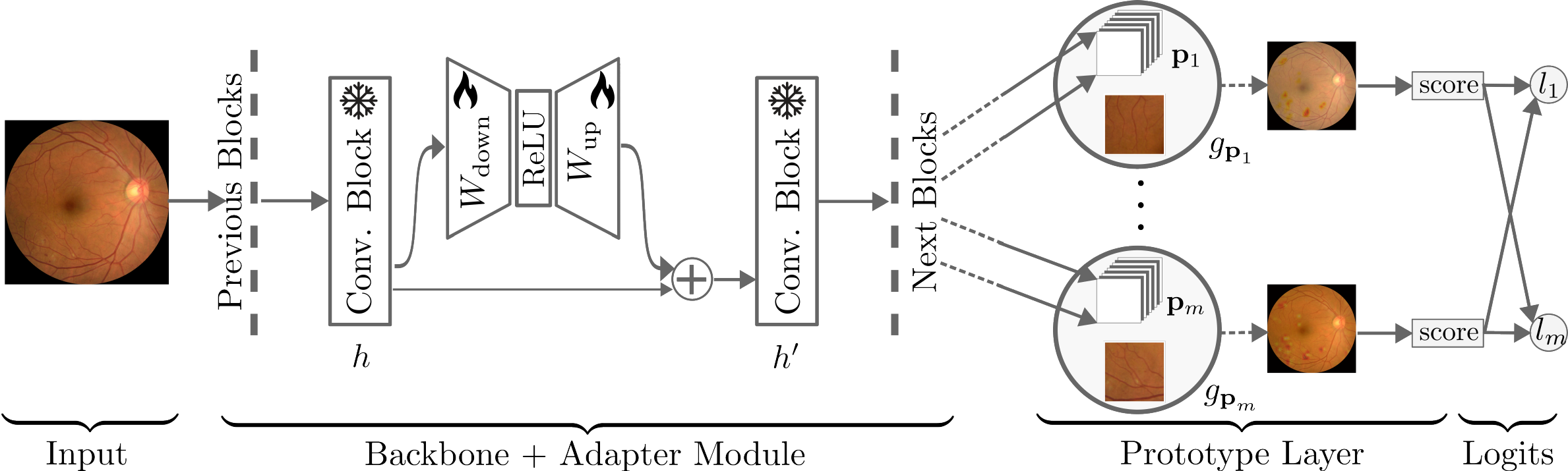}
		\caption{A depiction of the the model, in which an input image passes through a frozen backbone augmented with adapter modules. Each adapter module compresses a convolutional block, applies a non-linear activation (for example, ReLU), and expand it back, adding the result residually to the next convolutional block. After subsequent blocks, the output representation is compared to a set of learnable prototypes. Each prototype is convolved with the feature map to compute similarities, forming a score that is used to produce the final class logits.}
		\label{fig:model}
	\end{figure}
	\subsection{Background}
	In our federated learning (FL) setup, we utilise prototypical network~\cite{chen} with adapter modules as local models for the clients. We use adapters to act as compressed surrogates of the local models and guide clients to obtain generalisation regardless of varying client distribution. We attach adapter modules to different layers of the backbone model using a residual connection approach to prevent vanish gradient. The prototypes serve as learnable class-specific reference vectors that represent characteristic parts of different classes from the training dataset providing inherent interpretations and simultaneously classify the data. It is important to note that the backbone weight are always frozen and only the adapter modules and the prototype layers are optimised~(Figure~\ref{fig:model}).
	\paragraph{Adapters module.}
	The adapter module is a lightweight bottleneck network consisting of a down-sampling convolutional layer $W_{\text{down}} \in \mathbb{R}^{H' \times W' \times D \times r}$ and an up-sampling convolutional layer  $W_{\text{up}} \in \mathbb{R}^{H' \times W' \times r \times D}$, that is applied on a feature map $h \in \mathbb{R}^{H \times W \times D}$, where $r$ represents the down-sampled dimension and $D$ is the depth of the feature map $(r < D)$. We compute an adapter module as 	$h' = h + \sigma (h W_{\text{down}}) W_{\text{up}},$
	where $\sigma(\cdot)$ is a non-linear activation function and $h' \in \mathbb{R}^{H \times W \times D}$ is the resulting feature map. For the rest of the paper, we represent $W_{\text{down}}$ and $W_{\text{up}}$ as $\alpha$. 
	\paragraph{Prototype layer.}
	Suppose that $\mathbf{z} \in \mathbb{R}^{H \times W \times D}$ is the output of feature extractor. The prototype layer contains $m$ learnable prototypes $\mathbf{P} = \{\mathbf{p}_j\}_{j=1}^m$ with dimensions $\mathbf{p}_j \in \mathbb{R}^{h \times w \times D}$, where $1 \le h \le H, \, 1 \le w \le W$. Each prototype acts as a template matcher, computing similarity scores across all spatial positions in $\mathbf{z}$ having the same shape as $\mathbf{p}_j$ through the squared $\ell_2$ distance given as $d^{(h,w)}_j = \big\lVert \mathbf{z}^{(h, w)} - \mathbf{p}_j \big\rVert ^2_2$. In our case, we compute the similarity score by negating the distance. Each prototype $\mathbf{p}_j$ constitutes a learnable parameter $W_{\text{proto}} \in \mathbb{R}^{m \times h \times w \times D}$. For the rest of the paper, we represent $W_{\text{proto}}$ as $\phi$.
	\paragraph{Activation map and logits.}
	The similarity computations produce activation maps that preserve spatial information, indicating where in the image prototypical parts are detected. These can be upsampled to create heatmaps showing which image regions most strongly activate each prototype. Next, logits are obtained through a weighted combination of the similarity scores. This allows the model to both classify instances and provide inherent interpretations.
	\subsection{Problem setting}
	Let there be $N$ clients, indexed by $i \in \{1, \ldots N\}$. Each client $i$ has local data $\mathcal{D}_i = \{(x_j^{(i)}, y_j^{(i)})\}^{|\mathcal{D}_i|}_{j=1}$ of size $\mathcal{D}_i$. The total data is $|\mathcal{D}| = \sum_{i=1}^{N} |\mathcal{D}_i|$. In standard FedAvg~\cite{mcmahan}, the aim is to train a global model with parameters $\theta$. At each communication round $r \in \{1, \ldots, R\}$, the server broadcasts the current global model $\theta^r$ to each client $i$. Each client initialises its local model with the weights of the shared global model $(\theta_i^r \leftarrow \theta^r)$ and performs local updates for a number of epochs $E$ on local dataset $\mathcal{D}_i$. After local updates, the server collects local $\theta_i^r$ and updates the global model with $\theta^{r+1}$ obtained using a weighted average given by, $\theta^{r+1} = \sum_{i=1}^{N} \frac{|\mathcal{D}_i|}{|\mathcal{D}|} \theta_i^r$. FedAvg aims to minimise 
	\begin{equation}
		\underset{\theta}{\mathrm{min}} \sum_{i=1}^{N}  \frac{|\mathcal{D}_i|}{|\mathcal{D}|} \mathcal{L}_S(\mathcal{F}(\theta; x), y),
	\end{equation}
	where $\mathcal{L}_S$ is a loss function evaluated on data from client $i$ and $\mathcal{F}$ is the model parameterised by $\theta$. 
	\subsection{Proposed method}
	\begin{figure}[t]
		\centering
		\includegraphics[width=\textwidth]{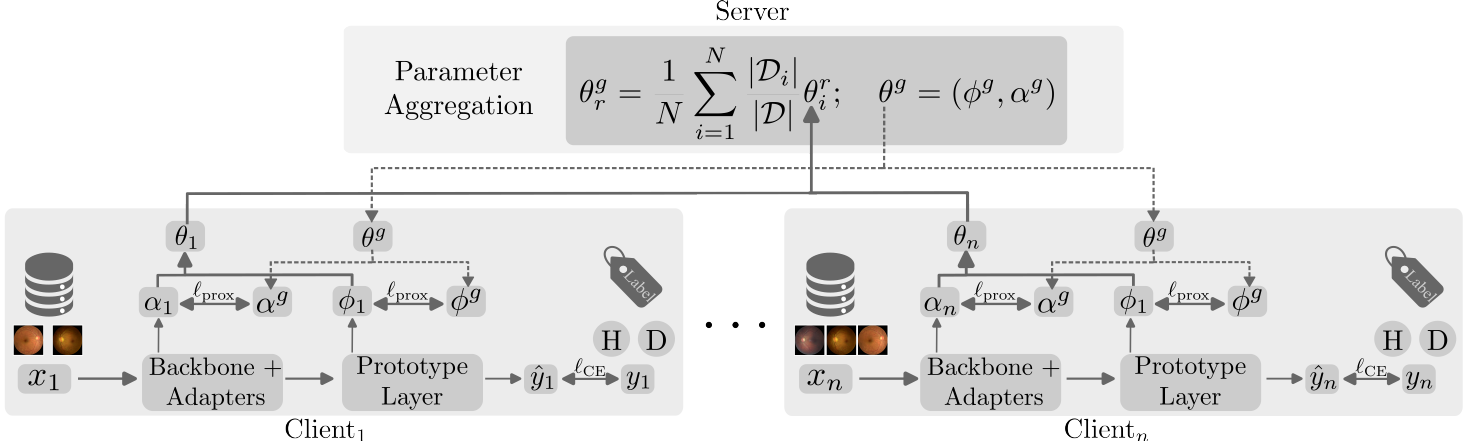}
		\caption{Illustration of our FL setup for adapter- and prototype-based learning. Each client $i$ receives from the server a shared global parameter set, comprising of adapter parameters and prototype parameters. Locally, the client refines the adapters and prototypes by minimising a combination of loss function and regularisers to align local parameters to the global ones. The updated parameters are sent back to the server, which performs parameter aggregation to obtain a new global model. This process is repeated for a number of rounds to enable each client to personlise adapters and prototypes for its local data distribution while collaboratively improving the global model.}
		\label{fig:workflow}
	\end{figure}
	We modify the above procedure to address statistical heterogeneity through adapters and prototypes. For this setup, each client has a prototypical part network (ProtoPNet) with three groups of parameters. For each client $i$, we have $\omega_i$ representing the backbone parameters, followed by the adapter module parameters denoted as $\alpha_i$, which are positioned at the end of every convolutional block of the backbone, and $\phi_i$ for the prototype parameters. We denote $\theta_i = (\alpha_i, \phi_i)$ as the communicating parameters from the clients to the server. We use the ResNet-50 architecture as our backbone in this work. Every ResNet model has four convolutional blocks that are made of several convolutional operations, thus our setup has four adapter modules. During training, we communicate only adapters and prototypes rather than the full model, improving communication overhead. Globally, we maintain a reference set of parameters $\theta^g$, containing prototypes $\phi^g$ and adapters $\alpha^g$, which are obtained via weight aggregation and sent back to the clients for update (Figure~\ref{fig:workflow}).
	\subsubsection{Local loss function} 
	Each client $i$ trains by minimising a combination of four main loss components. First, we compute the cross-entropy (CE) defined as $\ell_{\text{CE}} = - \sum_{c} \mathbf{1} \{y = c\} \log(\hat{y}_c)$ between the model's predicted logits $\hat{y}_c$ for class $c$ and the ground-truth labels $y$. Next for every client $i$, we use an $\ell_2$-norm to regularise the parameters of the adapter modules to prevent them from overfitting by computing $\ell_{\text{adapter}} = \beta \lVert \alpha_i \rVert_2^2$, where $\beta$ is a weighting coefficient for the adapter regulariser. Also, we regularise prototype by  encouraging a minimum distance $\ell_{\text{clst}}$, from each instance to correct-class prototype  and simultaneously encourage distance $\ell_{\text{sep}}$ to wrong-class prototype to be large. We apply an $\ell_1$ regularisation on the linear layer's prototype-to-class weights. Thus for every client $i$, the prototype regularisation is given by $\ell_{\text{proto}} = \lambda_{\text{clst}} \ell_{\text{clst}}(d) - \lambda_{\text{sep}} \ell_{\text{sep}}(d) + \gamma \lVert \phi_i \rVert_1$, where $\lambda_{\text{clst}}$, $\lambda_{\text{sep}}$ and $\gamma$ are weighting coefficients for the prototype regulariser. We align local adapter parameters $\alpha_i$ and prototype parameters $\phi_i$ to global references $\phi^g$ and $\alpha^g$ by adding a proximal penalty given by $\ell_{\text{prox}} = \frac{\mu_1}{2} \lVert \alpha_i - \alpha^g \rVert^2 + \frac{\mu_2}{2} \lVert\phi_i - \phi^g \rVert^2$, where $\mu_1$ and $\mu_2$ are coefficient that controls the strength of the regularisation. Putting it all together for each client $i$, we compute a local loss given as $\ell_i(\theta_i; x, y) = \ell_{\text{CE}}(y, \hat{y}) + \ell_{\text{proto}}(d) + \ell_{\text{adapter}}(\alpha_i) + \ell_{\text{prox}}(\phi_i,\alpha_i; \phi^g, \alpha^g)$.
	\subsubsection{Local update procedure and global aggregation}
	Within each communication round $r$, client $i$ receives $\theta_i^r = (\phi_{r - 1}^g, \alpha_{r - 1}^g)$ from the server. Each clients updates for a number of epochs iterating $\theta_i \leftarrow \theta_i - \eta \nabla_{\theta_i} \ell_i(\theta_i; x, y),$ where $\eta$ is the learning rate. After updates, client $i$ communicates $\phi_i$ and $\alpha_i$ back to the server. For every communication round $r$, we perform global update by averaging the parameters $\theta_i$ using the expression $\theta_r^g = \frac{1}{N} \sum_{i=1}^{N} \frac{\vert \mathcal{D}_i \vert}{\vert \mathcal{D} \vert} \theta_i^r$.
	\section{Experimental details}
	\paragraph{Datasets.}
	We conducted our experiments on retinal fundus images from EyePACS~\cite{eyepacs}. For this study, we used images that showed both the macula and optic disk. In addition, we selected images labelled as "\textit{Excellent}", "\textit{Good}", and "\textit{Adequate}" in quality. The dataset contains clinical site information, which provided us with distribution differences for the study. From the available clinical site information, we focused on the five most frequent sites -- four of which served as distinct training client data (each split 80\%/20\% for training and validation) and one for testing. As a result, the dataset span unique patients and varying instance of images across the selected sites. Furthermore, we selected DR as our target for the study. Specifically, the five  sites contained $8{,}962$, $7{,}601$, $5{,}099$, $4{,}080$ and $3{,}104$ images, with $76.21\%$, $55.27\%$, $80.81\%$, $83.82\%$ and $81.51\%$ of them classified as healthy, respectively. We preprocessed the images by using a tightly centred-circle cropping technique~\cite{sarah} and resizing them to $512 \times 512$ dimension. In addition, we applied colour jitter transformations, random flips and random rotations to further augment the data.
	\paragraph{Implementation details.}
	We focus on DR as a binary task by classifying images as healthy or diseased. We utilised the ResNet-50 model~\cite{he} pretrained on ImageNet~\cite{deng} as the backbone and employed transfer learning approach across 100 communication rounds. Each local client was trained for one epoch before they were communicated to the server. We did this to prevent local clients from overfitting on their corresponding local data. We set the local batch size at 16 due to GPU memory constraints. We employed the Adam optimiser~\cite{kingma} at a learning of 0.0001 and balanced the data using a class-balanced sampler method~\cite{alhamoud}. We used accuracy as our performance metrics. All experiments were implemented using the PyTorch framework on an NVIDIA A100 GPU. 
	\section{Results and discussion}
	\subsection{Site-wise model performance}
	We presented classification accuracy results for four clients under different FL algorithms~(Table~\ref{tab:perf}). Notably, the FedAdapter model with proximal constraint performed best, achieving the highest average accuracy ($87.58\%$). By contrast, the FedAdapter without proximal constraint achieved moderate accuracy levels suggesting that its inclusion during local training can significantly enhance global performance. On the other hand, the prototype variants showed promising performance but generally fell short of the top-performing solutions. Although not the best performing algorithm, our proposed method ranks closely behind the FedAdapter only outperforming it by $0.23\%$ for Client 1. The results obtained by the proposed model demonstrates consistency among clients, suggesting that the proposed method can handle heterogeneous client data within a federated setting. This design therefore underlines the importance of merging local updates with global objectives, while preserving adaptation to their own data.
	\begin{table}[b]
		\centering
		\caption{Comparison of classification accuracy (in percentage) across four clients.}
		\label{tab:perf}
			\begin{tabular}{@{}lccccc@{}}
				\toprule
				\textbf{Algo} & \textbf{Client 1} & \textbf{Client 2} & \textbf{Client 3} & \textbf{Client 4} & \textbf{Avg. Acc} \\ \midrule
				FedAvg          & 78.82 & 66.59 & 65.08 & 67.57 & 69.52 \\
				FedProx         & 75.11 & 72.74 & 72.55 & 72.55 & 73.35 \\
				FedNova         & 76.82 & 76.92 & 76.41 & 76.51 & 76.66 \\
				FedAdapter *    & 65.57 & 71.61 & 74.82 & 76.06 & 72.02 \\
				FedAdapter      & 86.71 & 87.88 & 87.84 & 87.88 & 87.58  \\
				FedProto        & 59.21 & 60.69 & 61.63 & 61.45 & 60.75  \\
				FedPrototypical & 70.26 & 74.52 & 74.90 & 76.43 & 74.03  \\
				Ours            & 86.94 & 87.67 & 86.33 & 86.32 & 86.82 \\ \bottomrule
				\multicolumn{6}{l}{\small * indicate FedAdapter without proximal constraint.} \\
			\end{tabular}
	\end{table}
	\subsection{Explainability}
	\begin{figure}[t]
		\centering
		\includegraphics[width=\textwidth]{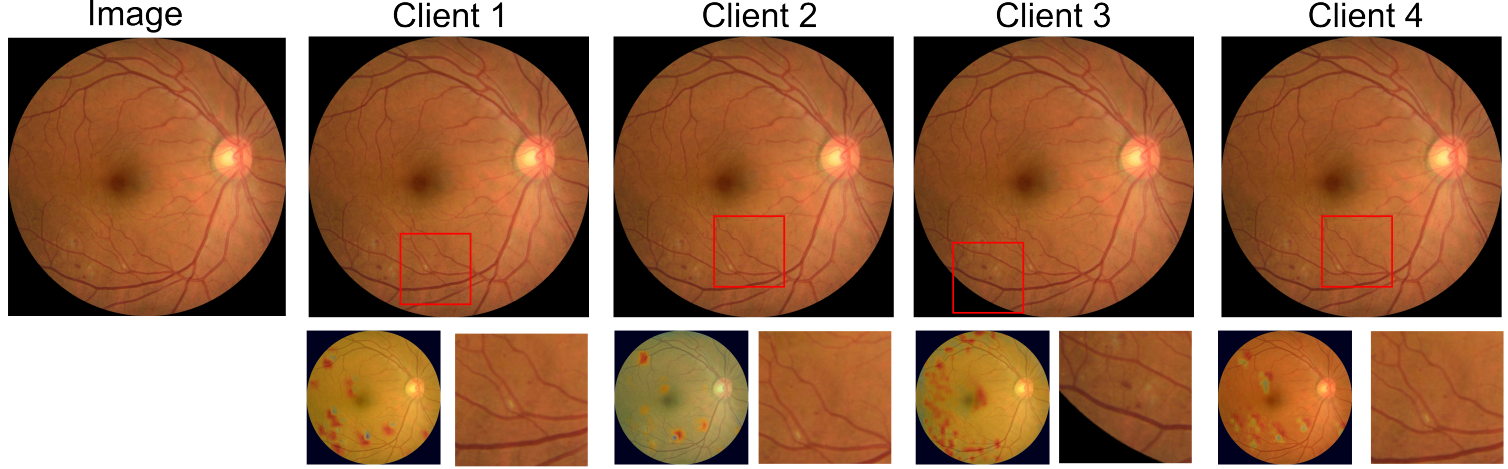}
		\caption{Visualising key disease retinal features from the proposed method. An input image is fed to the clients to obtain local prototypes highlighting disease markers on the retinal fundus images, indicated in red bounding box in the first row, along with their corresponding activation maps in the second row. The hotter colours on the activation map denotes positive activations where as cooler colours indicate negative activations. Also, the prototype part of the image is enlarged in the second row for clarity.}
		\label{fig:inter}
	\end{figure}
	Nonetheless, the benefits of our proposed method is apparent as it achieved good performance and provided inherent interpretation to model decisions. We visualised the prototypes by drawing a red bounding box at areas of the retinal fundus images where local prototypes fires strongly. In Figure~\ref{fig:inter}, we observed that all four clients placed their bounding box in almost the same area, showing a degree of agreement on what part of the image signals disease. This suggests that, even though each model was trained on a different dataset, they learned to focus on similar disease makers. We observed that most prototypes for DR detection mainly contained lesions-based features with normal or blood vessel background. These are typical features of DR. The activation maps also highlighted several hot regions on the retinal fundus image indicating possible regions of interests.
	\subsection{Performance on external dataset}
\begin{figure}[t]
		\centering
		\includegraphics[width=\textwidth]{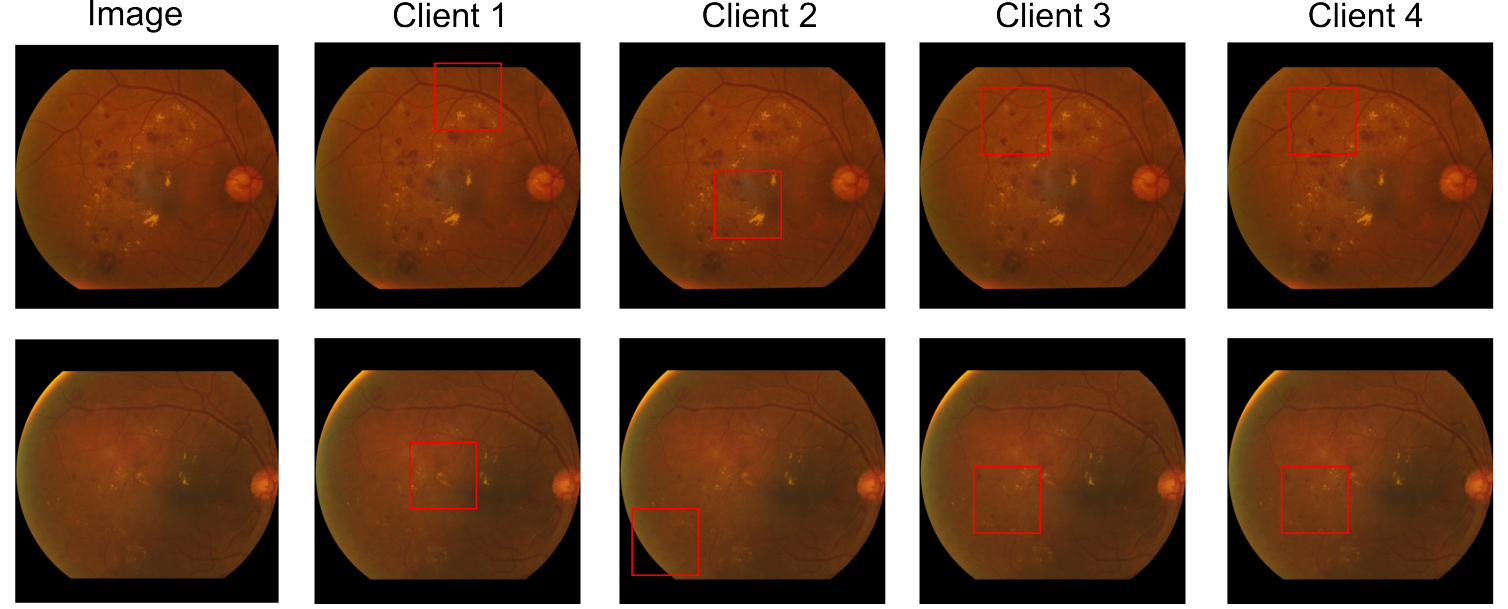}
		\caption{Demonstrating interpretability on external dataset with two randomly selected diseased images.}
		\label{fig:ext}
	\end{figure}

	We evaluated our method on the APTOS dataset to assess its generalisability. It contained $3660$ retinal fundus images with DR labels. We classified whether an image was healthy or diseased and obtained an accuracy of  $96.39\%$, $95.38\%$, $93.88\%$ and $95.57\%$ for clients 1, 2, 3 and 4 respectively. A key observation is that Clients 3 and 4 exhibit similar activation regions (Figure~\ref{fig:ext}). Their consistency could mean that they likely learned a common set of prototypical DR features that generalise well. In contrast, Clients 1 and 2 exhibit more variation in selected regions. 
	\section{Conclusion}
	We presented a novel FL framework that communicates only adapter modules and prototypes to help address statistical heterogeneity and provided inherent interpretations from the model. Regardless of the differences in distribution per client, the proposed method achieved consistent high accuracy scores and maintained interpretability by highlighting disease-relevant retinal regions across clients. Moreover, the ability of our proposed method to classify DR on an external data with high confidence suggests that it has learned core features of the disease and, thus, offer a promising implication for practical purposes, especially in healthcare. While the global diagnostic performance is robust, we observe that prototype-based explanations can differ between client models. In future, integrating model heterogeneity to the proposed model could address scenarios where clients have different network structure.
	\begin{credits}
		\subsubsection{\ackname} This project was supported by the Hertie Foundation and by the Deutsche Forschungsgemeinschaft under Germany’s Excellence Strategy with the Excellence Cluster 2064 “Machine Learning -- New Perspectives for Science”, project number 390727645. PB is a member of the Else Kröner Medical Scientist Kolleg “ClinbrAIn: Artificial Intelligence for Clinical Brain Research”. 
		\subsubsection{\discintname}
		The authors declare no competing interests.
	\end{credits}
	
\end{document}